\makeatletter\def\graphicscache@inhibit{true}\makeatother
\documentclass[letterpaper, 10 pt, conference]{ieeeconf}
\IEEEoverridecommandlockouts

\usepackage[utf8]{inputenc}

\usepackage{amsmath,amssymb,amsfonts}
\usepackage{graphicx}

\usepackage[hidelinks]{hyperref}
\usepackage[style=ieee,hyperref,natbib=true,backend=bibtex,firstinits,doi=false, %
     mincitenames=1,maxcitenames=2,maxbibnames=99,sorting=none,terseinits=false,hyperref=true]{biblatex}
\bibliography{references.bib}
\renewbibmacro*{bbx:savehash}{}%
\defbibheading{bibliography}[\bibname]{\section*{References}}

\usepackage{url}
\usepackage{graphicx}
\usepackage[usenames,dvipsnames]{xcolor}
\usepackage[draft]{fixme}
\fxsetup{theme=color}

\usepackage[capitalize]{cleveref}

\usepackage{tikz}
\usetikzlibrary{arrows}
\usetikzlibrary{arrows.meta}
\usetikzlibrary{positioning,calc}
\usetikzlibrary{decorations.pathreplacing}
\usetikzlibrary{decorations.markings}
\usetikzlibrary{fit}
\usetikzlibrary{shapes.callouts}
\usetikzlibrary{shapes.geometric}
\usetikzlibrary{matrix}

\usepackage{diagbox}
\usepackage{datetime}
\settimeformat{hhmmsstime}
\usepackage[binary-units=true,product-units=single,per-mode=symbol,range-units=single,range-phrase=\,--\,,detect-all]{siunitx} 
\DeclareSIUnit\dBFS{dBFS}

\usepackage[export]{adjustbox}

\usepackage{threeparttable}
\usepackage{booktabs}
\usepackage{multirow}

\IfFileExists{graphicscache.sty}{\usepackage{graphicscache}}

\makeatletter
\newcommand{\linebreakand}{%
  \end{@IEEEauthorhalign}
  \hfill\mbox{}\par
  \mbox{}\hfill\begin{@IEEEauthorhalign}
}
\makeatother

\begin{document}

\title{\LARGE \bf
Audio-based Roughness Sensing and Tactile Feedback \\ for Haptic Perception in Telepresence
}

\author{Bastian Pätzold$^{1*}$, Andre Rochow$^{1}$, Michael Schreiber$^{1}$, Raphael Memmesheimer$^{1}$, \\ Christian Lenz$^{1}$, Max Schwarz$^{1}$, and Sven Behnke$^{1}$%
\thanks{$^{1}$Autonomous Intelligent Systems, University of Bonn, Germany.}%
\thanks{$^{*}$Corresponding author. Email: {\tt paetzold@ais.uni-bonn.de}}%
}

\maketitle

\begin{abstract} %

Haptic perception is highly important for immersive teleoperation of robots, especially for accomplishing manipulation tasks.
We propose a low-cost haptic sensing and rendering system, which is capable of detecting and displaying surface roughness.
As the robot fingertip moves across a surface of interest, two microphones capture sound coupled directly through the fingertip and through the air, respectively.
A learning-based detector system analyzes the data in real time and gives roughness estimates with both high temporal resolution and low latency.
Finally, an audio-based vibrational actuator displays the result to the human operator.
We demonstrate the effectiveness of our system through lab experiments and our winning entry in the ANA Avatar XPRIZE competition finals, where briefly trained judges solved a roughness-based selection task even without additional vision feedback.
We publish our dataset used for training and evaluation together with our trained models to enable reproducibility of results.

Keywords: Haptics, Telepresence, Audio, Machine Learning
\end{abstract}

\section{Introduction}

Sensing surface properties through haptics is one of the fundamental ways, humans perceive their environment. 
Humans are able to perform a variety of exploratory movements with their hands and fingertips to discern aspects such as roughness, hardness, and shape of objects they manipulate~\citep{lederman1993extracting}.
It is widely understood that integrating haptics into VR, AR, and teleoperation systems is a key step towards increasing realism and acceptance of such systems~\citep{wee2021haptic}.
Consequently, numerous methods for sensing~\citep{fishel2012sensing,visuotactiles} and displaying~\citep{adilkhanov2022haptic} haptic sensations have been developed.
However, these systems are often highly complex, costly, and difficult to integrate into existing teleoperation systems, especially due to size restrictions.

In this work, we present the haptic system our team \mbox{NimbRo} developed for the ANA Avatar XPRIZE competition\addtocounter{footnote}{+1}\footnote{\url{https://www.xprize.org/prizes/avatar}}~\citep{schwarz2021nimbro,schwarz2023nimbro}.
The competition focused on intuitive and immersive telepresence in a mobile robot, including social interaction as well as manipulation capabilities.
To evaluate the intuitiveness of the developed telepresence systems, briefly trained members of the judging panel had to solve through them a sequence of ten increasingly difficult tasks.
The last and most difficult task focused on haptic perception, challenging the operators to discern two types of stones based on their surface roughness,
i.e. ``Was the Avatar able to feel the texture of the object without seeing it, and retrieve the requested one?''.

\begin{figure}
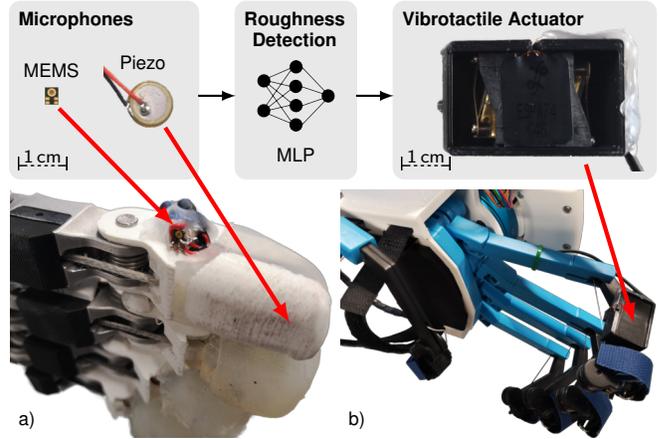

     \centering
     \begin{tikzpicture}[a/.style={inner sep=0pt}, l/.style={anchor=south west, fill=white, font=\sffamily\scriptsize}, b/.style={anchor=south west, draw, rounded corners, align=center, fill=white, font=\sffamily\scriptsize}]
       \newlength{\fingerheight}\setlength{\fingerheight}{.38\linewidth}
       \definecolor{background1}{rgb}    {0.91,0.91,0.91}
       \definecolor{background2}{rgb}    {0.39,0.73,0.83}
       \definecolor{orange}{rgb}         {1.00,0.69,0.00}
       
       \def\boxhgap{0.5};
       \def\boxvgap{0.15};
       \def\boxheight{2.5};
       \def\arrowheight{1.25};    
       \def\boxalen{2.5};
       \def\boxblen{1.6};
       
       \node[a] (img1) {\includegraphics[height=\fingerheight]{images/hardware/finger.png}};
       \node[a, right=.1cm of img1] (img2) {\includegraphics[height=\fingerheight]{images/hardware/glove.png}};
       \node[l] at ($(img1.south west)-(0.01,0.01)$) {a)};
       \node[l] at ($(img2.south west)-(0.01,0.01)$) {b)};
   
       \draw[thick, background1, rounded corners, fill] ($(img1.north west)+(0.0,\boxvgap)$) -- ($(img1.north west)+(\boxalen,\boxvgap)$) -- ($(img1.north west)+(\boxalen,\boxheight)$) -- ($(img1.north west)+(0.0,\boxheight)$) -- cycle;
       \node[anchor = north west, font=\sffamily\scriptsize] at ($(img1.north west)+(0.0,\boxheight)-(0.01,0.01)$) {\textbf{Microphones}};
   
       \draw[thick, background1, rounded corners, fill] ($(img1.north west)+(\boxalen+\boxhgap,\boxvgap)$) -- ($(img1.north west)+(\boxalen+\boxhgap+\boxblen,\boxvgap)$) -- ($(img1.north west)+(\boxalen+\boxhgap+\boxblen,\boxheight)$) -- ($(img1.north west)+(\boxalen+\boxhgap,\boxheight)$) -- cycle;
       \node[anchor = north west, align=center ,font=\sffamily\scriptsize] at ($(img1.north west)+(0.0+\boxalen+\boxhgap,\boxheight)-(0.01,0.01)$) {\textbf{Roughness}\\\textbf{Detection}};
   
       \draw[thick, background1, rounded corners, fill] ($(img1.north west)+(\boxalen+\boxhgap+\boxblen+\boxhgap,\boxvgap)$) -- ($(img2.north east)+(0.0,\boxvgap)$) -- ($(img2.north east)+(0.0,\boxheight)$) -- ($(img1.north west)+(\boxalen+\boxhgap+\boxblen+\boxhgap,\boxheight)$) -- cycle;
       \node[anchor = north west, font=\sffamily\scriptsize] at ($(img1.north west)+(\boxalen+\boxhgap+\boxblen+\boxhgap,\boxheight)-(0.01,0.01)$) {\textbf{Vibrotactile Actuator}};
   
       \draw[thick,-latex] ($(img1.north west)+(\boxalen,\arrowheight)$) -- ($(img1.north west)+(\boxalen+\boxhgap,\arrowheight)$);
       \draw[thick,-latex] ($(img1.north west)+(\boxalen+\boxhgap+\boxblen,\arrowheight)$) -- ($(img1.north west)+(\boxalen+\boxhgap+\boxblen+\boxhgap,\arrowheight)$);
   
       \node(h1)[a, anchor=center] at ($(img1.north west)+(0.55,\arrowheight)$) {\includegraphics[scale=0.08]{images/hardware/hardware_MEMS.png}};
       \node[a, right=0.6cm of h1] (h2) {\includegraphics[scale=0.08]{images/hardware/hardware_piezo.png}};
       \node(h4)[a, anchor=east] at ($(img2.north east)+(-0.15,1.20)$) {\includegraphics[scale=0.08]{images/hardware/hardware_basslet.png}};
       
       \node(scale_a)[a, anchor=south west] at ($(img1.north west)+(0.0,\boxvgap)+(0.1,0.1)$) {\includegraphics[scale=0.08]{images/hardware/hardware_10mm.png}};
       \node[font=\sffamily\scriptsize, above=-0.2cm of scale_a] {$\SI{1}{\cm}$};
   
       \node(scale_b)[a, anchor=south west] at ($(img1.north west)+(\boxalen+\boxhgap+\boxblen+\boxhgap,\boxvgap)+(0.1,0.1)$) {\includegraphics[scale=0.08]{images/hardware/hardware_10mm.png}};
       \node[font=\sffamily\scriptsize, above=-0.2cm of scale_b] {$\SI{1}{\cm}$};
       
       \node(h1l)[anchor=south, font=\sffamily\scriptsize] at ($(h1.north)+(0.0,0.0)$) {MEMS};
       \node(h2l)[anchor=south, font=\sffamily\scriptsize] at ($(h2.north)+(0.05,-0.2)$) {Piezo};
   
       \node(l1) at ($(img1.north west)+(2.3,-0.6)$) {};
       \draw[-latex,ultra thick,color=red] ($(h1.south)+(+0.1,-0.05)$) -- (l1);
   
       \node(l2) at ($(img1.north west)+(3.8,-1.9)$) {};
       \draw[-latex,ultra thick,color=red] ($(h2.south)+(0.35,-0.0)$) -- (l2);
   
       \node(l3) at ($(img2.north east)-(0.25,1.9)$) {};
       \draw[-latex,ultra thick,color=red] ($(h4.south)+(0.6,0.1)$) -- (l3);
   
       \node(mlp)[a, anchor=center] at ($(img1.north west)+(\boxalen+\boxhgap+0.8,\arrowheight)$) {\includegraphics[scale=0.03]{images/hardware/MLP.png}};
       \node[font=\sffamily\scriptsize, below=0.1cm of mlp] {MLP};
       
     \end{tikzpicture}
     \caption{Hardware implementation for roughness sensing and tactile feedback integrated with our telepresence system. Instrumented index fingers on (a)~Schunk SIH robot hand and (b) SenseGlove DK1 hand exoskeleton.}
     \vspace{-0.48\baselineskip}
     \label{fig:hardware}
\end{figure}

In contrast to previous works, our proposed haptic sensing and display system achieves roughness sensing at very low cost by using off-the-shelf audio components.
Both sensing and display components are compact and easily integrated into teleoperation systems as exemplified in \Cref{fig:hardware}.

Our approach is based on capturing audio signals using two different types of microphones. These audio signals are analyzed by a neural network trained on a custom dataset providing exemplary surface contacts with various stones and other objects. 
The operator is notified about the presence and roughness of the perceived surface through low-latency vibratory feedback, which aims to convey an intuitive sense of touch that does not require special training of the operator.

The system was successfully evaluated at the ANA Avatar XPRIZE finals, where three different operator judges solved the haptic perception task as well as all other tasks in the fastest time, winning our team NimbRo the \$5M grand prize.

In summary, our contributions include:
\begin{enumerate}
     \item a compact and low-cost hardware design of both sensing and display components,
     \item a learning-based method for online, low-latency and high temporal resolution roughness analysis, and
     \item an evaluation of this system in the competition as well as in fully reproducible offline experiments.
\end{enumerate}

\begin{figure*}
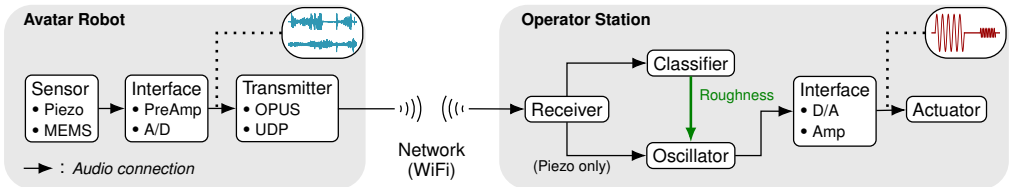

     \centering
     \resizebox{0.78\textwidth}{!}{%
\begin{tikzpicture}[font=\sffamily,
arr/.style={-{Latex[length=0.8mm]}},
transmission/.style={decorate, decoration={expanding waves, angle=30,
                          segment length=1}},
very thin
] \footnotesize

\definecolor{background1}{rgb}    {1.00,1.00,1.00} %
\definecolor{background2}{rgb}    {0.91,0.91,0.91} %
\definecolor{background3}{rgb}    {0.99,0.99,0.99} %
\definecolor{orange}{rgb}         {0.56,0.7,0.24} %

\def\gap{0.1}
\def\middle{0.420}
\def\width{5.0}
\def\height{1.17}

\draw[thick, rounded corners, background2,fill] (\gap, \gap) -- (\width*\middle-\gap*0.75, \gap) -- (\width*\middle-\gap*0.75, \height-\gap) -- (\gap, \height-\gap) -- cycle;

\def\labelwidth{1}
\def\labelheight{10}
\def\labelscale{0.35}
\def\thumb_scale{0.15}

\def\labeloffsety{\height-\gap*4-0.03}

\node[scale=0.9*\labelscale, anchor=north west] at (\gap*2 - 0.04, \height-\gap) {\textbf{Avatar Robot}};

\tikzset{content_node/.append style={minimum size=0,minimum height=\labelheight,minimum width=\labelwidth,draw,align=left,scale=\labelscale,fill=white,rounded corners=1pt}}

\node(Sensor)[content_node, anchor=north west] at (\gap*2,\labeloffsety-0.035) {Sensor\\{\raise .32ex\hbox{\tiny$\bullet$}}\scriptsize~Piezo\\{\raise .32ex\hbox{\tiny$\bullet$}}\scriptsize~MEMS}; 

\node(Interface1)[content_node, anchor=north west] at (\gap*2+ 0.54,\labeloffsety-0.035) {Interface\\{\raise .32ex\hbox{\tiny$\bullet$}}\scriptsize~PreAmp\\{\raise .32ex\hbox{\tiny$\bullet$}}\scriptsize~A/D};
\node(Transmitter)[content_node, anchor=north west] at (\gap*2+ 1.14 ,\labeloffsety-0.035) {Transmitter\\{\raise .32ex\hbox{\tiny$\bullet$}}\scriptsize~OPUS\\{\raise .32ex\hbox{\tiny$\bullet$}}\scriptsize~UDP};

\begin{scope}[shift={(0.6,0)}]
    \draw[thick, rounded corners, background2,fill] (\width*\middle+\gap*0.75, \gap) -- (\width-\gap, \gap) -- (\width-\gap, \height-\gap) -- (\width*\middle+\gap*0.75, \height-\gap) -- cycle;
    \node[scale=0.9*\labelscale, anchor=north west] at (\gap*2 + 2.05, \height-\gap) {\textbf{Operator Station}};
    \node(Receiver)[content_node, anchor=north west] at (\gap*2+ 2.1 ,\labeloffsety - 0.15) {Receiver};
    \node(Classifier)[content_node, anchor=north west] at (\gap*2+ 2.76 ,\labeloffsety + 0.10) {Classifier};
    \node(Oscillator)[content_node, anchor=north west] at (\gap*2+ 2.757 ,\labeloffsety - 0.40) {Oscillator};
    \node(Interface2)[content_node, anchor=north west] at (\gap*2+ 3.55 ,\labeloffsety - 0.042) {Interface\\{\raise .32ex\hbox{\tiny$\bullet$}}\scriptsize~D/A\\{\raise .32ex\hbox{\tiny$\bullet$}}\scriptsize~Amp};
    \node(Display)[content_node, anchor=north west] at (\gap*2+ 4.16 ,\labeloffsety - 0.15) {Actuator};
    \node[scale=1.3*\labelscale, anchor=north west] at (\gap*2+ 2.1 ,\labeloffsety - 0.45) {\tiny (Piezo only)};
    \node(thumb_osc)[content_node,fill=white,rounded corners=.12cm, draw=black, anchor=north east] at (\width-\gap,\height-\gap) {\includegraphics[scale=\thumb_scale]{figures/pipeline/thumbnail_osc.pdf}};
\end{scope}

\draw[arr] (Sensor) -- (Interface1);
\draw[arr] (Interface1) -- (Transmitter);

\draw[arr] (Receiver.north) |- (Classifier.west);
\draw[arr] (Receiver.south) |- (Oscillator.west); %
\draw[arr,line width=0.4,color=green!50!black] (Classifier) -- (Oscillator) node[pos=0.3,right,scale=1.3*\labelscale]{\tiny Roughness};

\draw[arr] (Oscillator.east) -| node[scale=\labelscale] {} ++(0.14,0.24) -- (Interface2.west);
\draw[arr] (Interface2) -- (Display);

\node(ArrowStart)[anchor=center] at (\gap-0.005 ,2*\gap) {};
\node(ArrowStop)[anchor=center,scale=\labelscale] at (\gap*4,2*\gap) {:};
\node(ArrowLabel)[anchor=west ,scale=\labelscale] at (\gap*4.2,2*\gap) {\scriptsize \textit{Audio connection}};
\draw[arr] (ArrowStart) -- (ArrowStop);

\node(thumb_mics)[content_node,fill=white,rounded corners=.12cm, draw=black, anchor=north east] at (\width*\middle-\gap*0.75,\height-\gap) {\includegraphics[scale=\thumb_scale]{figures/pipeline/thumbnail_mics.pdf}};

\draw[thin,densely dotted,color=black] (thumb_mics.west) -| node[scale=\labelscale] {} ++(-0.3512 ,0) -- node[scale=\labelscale] {} ++(0 ,-0.4);
\draw[thin,densely dotted,color=black] (thumb_osc.west) -| node[scale=\labelscale] {} ++(-0.204 ,0) -- node[scale=\labelscale] {} ++(0 ,-0.4);

\coordinate (wifi) at ($(Transmitter.east)!0.5!(Receiver.west)$);
\coordinate (wifi_start) at ($(wifi)+(-0.2,0)$);
\coordinate (wifi_stop) at ($(wifi)+(0.2,0)$);
\draw[transmission,very thin] (wifi_start) -- ($(wifi)+(-0.05,0)$);
\draw[transmission,very thin] (wifi_stop) -- ($(wifi)+(0.05,0)$);

\draw (Transmitter.east) |- (wifi_start);
\draw[arr] (wifi_stop) -- (Receiver);

\node[below=0.15 of wifi,scale=\labelscale,align=center] {Network\\(WiFi)};

\end{tikzpicture}
}
     \caption{Proposed end-to-end pipeline for audio-based roughness sensing and tactile feedback in telepresence applications.}
     \label{fig:pipeline}
\end{figure*}

\pagebreak

\section{Related Work}

\subsection{Tactile Sensing}

Tactile sensors are based on a wide range of sensing principles, including capacitance, resistance, pressure, magnetism, and optics. %
For example, \citet{fishel2012sensing} introduced the \textit{BioTac} tactile sensor, based on an incompressible liquid as an acoustic conductor.
In addition to its capability of measuring shear forces, skin stretch and temperature, it detects vibrations with up to $\SI{1040}{\Hz}$ using a pressure sensor.
By using only a single sensor per fingertip, it has a low spatial resolution, though.
\textit{GelSight}, proposed by \citet{yuan2017gelsight}, is capable of measuring high-resolution geometry as well as local and shear forces by visually observing the deformation of an elastomer sensor surface with an embedded camera.

Despite the promising capabilities of such devices, they suffer from two disadvantages from our point of view:
First, their size might be considered too large for integration in existing hardware solutions, or deployment in large quantities with high spatial resolution.
Second, their availability and high cost limit their feasibility for numerous applications.

Our work focuses on deploying considerably smaller and lower-cost audio-based hardware.
In a similar manner, \citet{tires} describe the utilization of microphones to classify road surfaces by capturing the tire-pavement interaction noise in an automotive context.
They convert the audio signals to time-frequency RGB images and feed them to a CNN.
Even though the captured audio signals depend on other factors besides the road surface, such as the car speed, tire type, and wheel torque, they demonstrate the effectiveness of their approach for classifying snow and asphalt surfaces. 
\citet{piezo} use a piezo acoustical sensor to analyze irregularities in materials, such as aluminum or stainless steel, occurring during manufacturing.
They capture the friction sounds when moving a stylus with a diamond tip over the surface of a specimen.
The controlledness of the environment allows the determination of roughness parameters~\cite{iso} using classical signal processing approaches.
Microphones have also been used to identify touch and swipe gestures on mobile devices~\citep{sun2018vskin,lopes2011augmenting}.

Most similar to our use case, \citet{svensson2021electrotactile} help prosthetic users feel textures
by stroking a microphone across object surfaces. In contrast to our system, the signal
is filtered in a fixed manner, extracting the median frequency, which is then applied to the
user using electrostimulation. This limits the scope to regular textures (such as mesh, rubber, etc.),
where a frequency is easily extracted.
In contrast, our method works on irregular surfaces such as natural stones.

\subsection{Tactile Rendering}

Tactile actuators are integrated into numerous devices, such as phones or game controllers.
Due to the rising interest in telepresence and VR applications, a wide range of haptic displays -- typically referred to as \textit{Haptic Gloves} -- emerged in recent years~\citep{glovesurvey2,glovesurvey1}, promising to convey realistic kinesthetic and tactile feedback.
Typically, tactile feedback is achieved by displaying vibrations utilizing eccentric rotating masses, linear resonant actuators, or piezoelectric actuators.
Their limitations often include support for only a single resonant frequency, poor intensity resolution, and slow response times.
Instead, we utilize an acoustic actuator to address these issues while maintaining comparable size and costs.

\section{Method}

In this section, we describe our method for sensing and actuation in detail. \Cref{fig:pipeline} gives an overview of the pipeline.

\subsection{Sensing}\label{sec:sensing}

The surface point of the avatar robot to be provided with roughness sensing capabilities (e.g. the tip of an index finger) is equipped with two microphones (\cref{fig:finger_cad}).
A piezo microphone is attached directly to the inside of the chosen surface to measure vibrations within the robot structure,
while a MEMS-type microphone is placed in close proximity ($\sim\SI{2}{\cm}$) to the outside of the chosen surface measuring vibrations in the air around it.
Once the surface makes contact with and slides over the unknown texture of an object, a sound gets induced into both microphones. %
These audio signals are then leveraged to classify the unknown texture as either \textit{smooth} or \textit{rough}.

\subsection{Classical Detection}

Initially, we attempted to find a direct mapping between the piezo microphone and our haptic display, utilizing classical approaches like dynamics processors and filters.
While it seemed easy for our team to classify the considered textures by directly hearing the piezo microphone signal, we could not find a suitable transformation to accommodate our haptic perception and the properties of the considered actuator.
Our failed attempts focused on isolating frequency regions critical for this perception, enhancing their transients, and pitch-shifting them into lower registers supported by the actuator. %

Next, we decoupled the classification from the signal sent to the actuator, by generating a new audio signal using a sine oscillator that conveys the haptic perception associated with the classification result.
While we found the utilization of an oscillator with varying frequency and amplitude to be intuitive and convincing for conveying different textures, the approach to classification was not satisfactory for our application.
In general, we found that rough textures induce louder and more transient signals into the piezo microphone than smooth textures, but the ambiguous amount of pressure and speed applied by the operator mask these effects.
Likewise, a hand-held solid stone induces a signal that differs strongly in level and frequency spectrum from a hollow stone mounted inside a box, although their textures are very similar.

In summary, while we found a functioning configuration for a limited number of objects and scenarios, the classical approach lacked generalization across situations and users.

\subsection{Learned Detection}

Instead of hand-designed filters, we opted for a learning-based approach.
As the teleoperation task demands low-latency haptic feedback, we update the prediction with every received audio buffer ($\sim\SI{10}{\ms}$) by constructing chunks that have access to $\SI{256}{\ms}$ of the past.
After low-pass filtering and reducing the sampling frequency, we calculate the FFT and concatenate the norm of both signals.
Experiments showed a sampling frequency of $\SI{2}{\kHz}$ to be sufficient for representing the relevant features for the described task.
Classification is then performed by an MLP with 15 hidden layers of which ten layers, with $256$ hidden units each, are equipped with residual connections for a better gradient flow. %
Experiments have shown that the classification accuracy is increased when the unnormalized input of both microphones is used, maintaining the relative loudness differences between them. %
When sliding over smooth surfaces -- in contrast to rough ones -- the MEMS microphone's level tends to be significantly quieter than that of the piezo microphone.
Such patterns can easily be learned by a neural network and should therefore not be discarded through normalization.
In fact, they constitute our motivation for deploying the additional MEMS microphone.

During inference, we detect the loudness of the piezo microphone in real time and compare it against a preset threshold that slightly exceeds its noise floor.
This allows to distinguish between \textit{contact} and \textit{no~contact} situations, which is used to gate the classification output of the network.

\subsection{Actuation}

The classification results are used to update the amplitude and frequency of a simple sine oscillator.
In the case of a \textit{smooth} result, we set it to a low amplitude and a high frequency (e.g. $\SI{120}{\hertz}$),
while for a \textit{rough} result, we set a higher amplitude and a lower frequency (e.g. $\SI{60}{\hertz}$), aiming to convey the feel of the texture, respectively.
Both parameters are low-pass filtered to produce a smooth waveform.
For the \textit{no~contact} case, the amplitude is set to zero. 

The generated audio signal is sent to a compact loudspeaker with a special voice-coil design, capable of reproducing frequencies perceivable by the human skin (${<\SI{1}{\kilo\hertz}}$)~\cite{bolanowski1988four} in the form of intense but mostly inaudible vibrations. %
The speaker is attached to the operator station matching the sensor position on the avatar robot (e.g. the fingertip of a hand exoskeleton).

The latency of the end-to-end haptic feedback is defined by the chunk size of the classification network, the buffer size of the avatar-side and operator-side audio systems, as well as the network transmission latency between both systems.

The haptic feedback allows the operator to intuitively distinguish between the situations: \textit{no~contact}, \textit{contact~with~a~smooth~texture}, and \textit{contact~with~a~rough~texture}.
The operator's haptic perception of smooth textures can be described as \textit{fizzy}, while the perception of rough textures might be described as \textit{bumpy}.
The high temporal resolution of the ternary classifier allows the operator to estimate the degree of roughness and to identify local irregularities.

\section{Implementation} %

\begin{figure}[t]
     \centering
     \resizebox{0.5\textwidth}{!}{%
\begin{tikzpicture}[font=\sffamily] \footnotesize

\def\labelwidth{1}
\def\labelheight{10}
\def\labelscale{1}

\tikzset{label/.append style={minimum size=0,minimum height=\labelheight,minimum width=\labelwidth,align=left,scale=\labelscale}}

\node(finger)[anchor=south east] at (0,0) {\includegraphics[width=.96\linewidth]{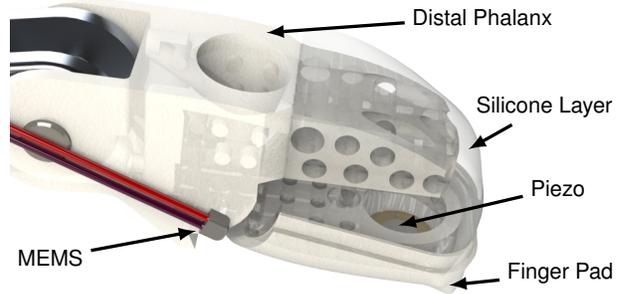}};

\node(distal)[label, anchor=west] at (-3.32,3.7) {Distal Phalanx};
\node(silicone)[label, anchor=west] at (-2.5,2.55) {Silicone Layer};
\node(finger_pad)[label, anchor=west] at (-2.1,0.4) {Finger Pad};

\node(piezo)[label, anchor=west] at (-1.8,1.5) {Piezo};
\node(mems)[label, anchor=west] at (-8.4,0.6) {MEMS};

\draw[-latex,very thick] (distal) -- (-5.1, 3.5);
\draw[-latex,very thick] (silicone) -- (-2.5, 2.0);
\draw[-latex,very thick] (piezo) -- (-3.4, 1.02);
\draw[-latex,very thick] (finger_pad) -- (-2.75, 0.25);
\draw[-latex,very thick] (mems) -- (-6.0, 0.95);

\end{tikzpicture}
}
     \caption{CAD drawing of sensorized Schunk SIH index finger. The MEMS and piezo microphones are glued to the 3D-printed distal phalanx and finger pad. A silicone layer connects both components.}
     \label{fig:finger_cad}
\end{figure}

\subsection{Avatar Robot}

Both microphones are attached to the left index finger of our avatar robot's \textit{Schunk SIH} hand (\cref{fig:hardware}).
We replaced the original index finger with a 3D-printed distal phalanx and finger pad (\cref{fig:finger_cad}). 
Both feature holes which allow a silicone layer to connect them.
The piezo microphone is glued to the inside surface of the finger pad.
The MEMS microphone is placed on the side of the finger, where it is close to the fingertip, but does not interfere during manipulation tasks.
To avoid the proximity effect affecting the network's classification performance, we chose a MEMS microphone that is omnidirectional and thus does not exhibit proximity.
The silicone layer is slightly compliant and decouples the finger pad from the rest of the robot, preventing vibrations to spill over into the piezo microphone.
The finger pad shape is designed to allow for sliding over a wide range of textures without getting stuck, while also producing a suitable amount of vibrations in the finger to allow for reliable classification.

The microphones are connected to a \textit{Focusrite~Scarlett~2i2} interface, which is used for pre-amplification and A/D conversion.
We set both inputs to \textit{high~impedance} mode, which maximizes the microphones' frequency response.
The digital audio signals are forwarded and processed using the \textit{JACK~Audio~Connection~Kit} which operates on top of the \textit{Advanced~Linux~Sound~Architecture}.
Both signals are transmitted from the avatar robot to the operator station by a low-latency UDP transmitter utilizing the \textit{OPUS~audio~codec}~\cite{opus}.

\subsection{Operator Station}

The operator station receives the audio signals of both microphones within a similar JACK environment as described for the avatar robot.
Both signals are forwarded to the classification network as described in \cref{sec:sensing}. %
The confidence of the classifier modulates the frequency and amplitude of a sine oscillator.
A \textit{rough} classification targets a frequency of $\SI{60}{\hertz}$ and a level of $\SI{0}{\dBFS}$.
Correspondingly, a \textit{smooth} classification targets a frequency of $\SI{120}{\hertz}$ and a level of $\SI{-25}{\dBFS}$.
Finally, we measure the loudness of the signal originating from the piezo microphone to discern \textit{contact} and \textit{no~contact} situations. 
The low noise floor and mechanical decoupling of the piezo microphone allow us to easily find a suitable fixed threshold parameter to facilitate a reliable and sensitive way of measuring contact.
A \textit{no~contact} situation then overwrites the roughness classifier and modulates the amplitude of the oscillator to a target level of $-\infty\,\SI{}{\dBFS}$.
We low-pass filter amplitude and frequency modulations to prevent artifacts in the generated waveform arising from altering classification results.

The generated audio signal is forwarded to our vibrotactile actuator shown in \cref{fig:hardware}.
It has been extracted from a \textit{Lofelt~Basslet}, a wearable consumer device designed to provide the sensation of bass when listening to music through headphones.
It offers fast acceleration response across its frequency range of $\SI{35}{\hertz}$ to $\SI{1}{\kilo\hertz}$.
Originally, the device is designed to be wrist-worn, houses a rechargeable battery, and offers audio connectivity via Bluetooth.
Instead, we extract its vibrotactile actuator, embed it in a small-footprint 3D-printed case and drive it using the mainboard's onboard soundcard for D/A conversion and a \textit{Fosi~Audio~TP-02} subwoofer amplifier.
The case is attached to the left index finger of the \textit{SenseGlove~DK1} hand exoskeleton worn by the operator, inducing vibrations from above the tip.

As the chunk size processed by the MLP matches the buffer size of $512$ samples set on both, the avatar robot's and the operator station's audio systems running with a sampling frequency of $\SI{48}{\kHz}$, 
the latency of the entire audio system is $\SI{21}{\ms}$ (omitting further network transmission delays).

\subsection{Data Acquisition and Network Training}

\subsubsection{Dataset}

\begin{figure}
     \begin{tikzpicture}[a/.style={inner sep=0pt}, l/.style={anchor=south west, fill=white, font=\sffamily\scriptsize}]
          \node[a] (img) {\includegraphics[width=\columnwidth]{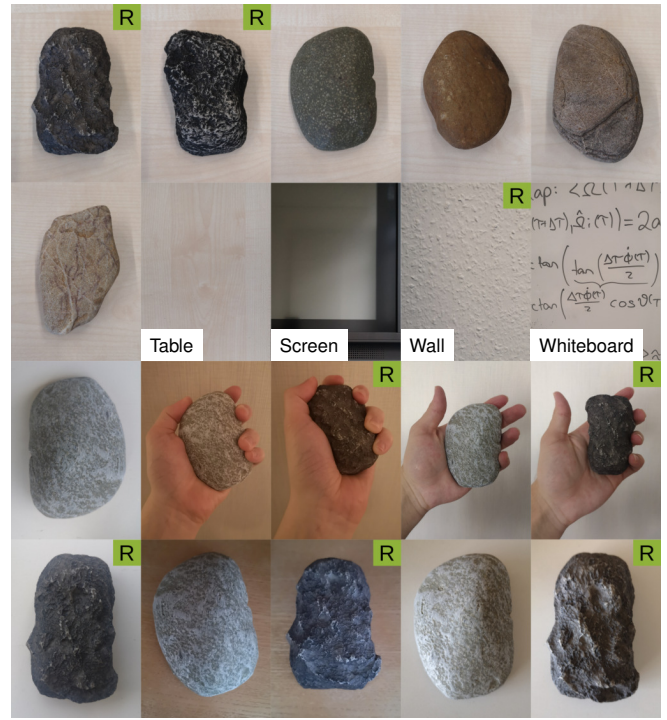}};
          \node[l] at (-2.6,-0.005) {Table};
          \node[l] at (-0.87,-0.005) {Screen};
          \node[l] at (0.855,-0.005) {Wall};
          \node[l] at (2.58,-0.005) {Whiteboard};
     \end{tikzpicture}
     \caption{Dataset objects for classification of rough and smooth textures. Rough object labels are indicated by green squares.}
     \label{fig:dataset}
\end{figure}

We recorded a custom dataset using our instrumented robot hand, making contact with and sliding over various textures with multiple patterns and intensities.
It consists of a training set with $20$ objects (\cref{fig:dataset}), and a test set with two objects (\cref{fig:textures}).
Each object is manually labeled as either \textit{rough} or \textit{smooth}. 
We deliberately chose objects where this distinction is explicit.
Since the task description in the Avatar XPRIZE competition finals clearly defined the requirement to classify the texture of artificial stones, we include various artificial stones in the dataset.
However, to improve generalization, we also include natural stones and other textures such as ingrain wallpaper and a wooden table surface.
Some objects are measured multiple times under varying acoustical scenarios (handheld, on a table, inside a box, etc.) to improve domain robustness.
For each object, we obtain seven recordings with a duration of $\SI{30}{\sec}$, including light, medium, and strong pressure levels, with long and short strokes each,
as well as a recording where we apply longer continuous wiggles, to support other interactive sensing approaches w.r.t. the operator.
As inference is performed on the operator station, we encode the training data using the \textit{OPUS~audio~codec}, mimicking transmission effects.

\subsubsection{Training}

We split the training data into chunks of $\SI{256}{\ms}$ and adopt the label of the respective file if the RMS loudness of the chunk exceeds a threshold determining a valid contact, similar to the threshold used to mute the oscillator output.
This ensures that all labeled chunks correspond to surface contacts, but conversely, not every contact is assigned to a labeled chunk.
Without consideration of these unlabeled chunks, the network would show unpredictable behavior at inference time.
Therefore, we introduce a third \textit{non-valid} class (not utilized during inference) which is comprised of chunks below the set threshold.
As the specific value of this threshold varies between experiments we report it in the evaluation.

The network is trained for five epochs with a batch size of $6000$ chunks.
We use a negative log-likelihood loss and the Adam optimizer with a learning rate of $1e$-$4$.
We add Gaussian noise to each chunk to prevent overfitting and improve generalizability.
This is particularly important, as external noises captured by the MEMS microphone or the audio circuitry might induce disturbances.

Both our dataset and the trained models are made public to enable reproducibility of results\footnote{\url{https://github.com/AIS-Bonn/Roughness_Sensing}}.

\section{Evaluation}

Our system has been evaluated in several steps, focusing on quantitative analysis of model training, as well as intuitiveness and immersion in a longer integrated mission.

\subsection{Quantitative Analysis}

\begin{table}
 \caption{General Model Confusion Matrices.}
 \label{tab:confusion}

 \begin{threeparttable}
 \begin{tabular}{@{}c@{\hspace{4pt}}crr}
  \multicolumn{4}{c}{Competition runs} \\
  \toprule
  & & \multicolumn{2}{c}{Response} \\
  \cmidrule(lr){3-4}
  &         & Rough & Smooth \\
  \midrule
  \multirow[c]{2}{*}{\rotatebox[origin=tr]{90}{Stone}}
  & Rough   & 11.3\,\% & 47.2\,\% \\[0.8ex]
  & Smooth  &  0.9\,\% & 40.5\,\% \\
  \bottomrule
 \end{tabular}
 \end{threeparttable}
 \hfill
 \begin{threeparttable}
 \begin{tabular}{@{}c@{\hspace{4pt}}crr}
  \multicolumn{4}{c}{Test set (\cref{fig:textures})} \\
  \toprule
  & & \multicolumn{2}{c}{Response} \\
  \cmidrule(lr){3-4}
  &         & Rough & Smooth \\
  \midrule
  \multirow[c]{2}{*}{\rotatebox[origin=tr]{90}{Stone}}
  & Rough   & 24.0\,\% & 26.5\,\% \\[0.8ex]
  & Smooth  &  8.1\,\% & 41.3\,\% \\
  \bottomrule
 \end{tabular}
 \end{threeparttable}

 \vspace{0.2cm}

 {\footnotesize The system is tuned to produce a low false-positive rate (i.e. smooth surfaces
 classified as rough). Results are obtained using the general model variant.}
\end{table}

\begin{table}
 \caption{Accuracy of Model Variants.}
 \label{tab:variants}
 \centering
 \setlength{\tabcolsep}{6pt}
 \begin{threeparttable}
 \begin{tabular}{l@{\hspace{7.5pt}}rrrrrr}
  \toprule
        & \multicolumn{4}{c}{Competition runs}  \\
  \cmidrule (lr) {2-5}
        & \multicolumn{2}{c}{Rough} & \multicolumn{2}{c}{Smooth} & \multicolumn{2}{c}{Test set} \\
  \cmidrule (lr) {2-3}
  \cmidrule (lr) {4-5} \cmidrule (lr) {6-7}
  Model & Day 1 & Day 2 & Day 1 & Day 2 & Rough & Smooth \\
  \midrule
  General    & \textbf{0.264} & 0.141          & \textbf{1.000} & 0.929          & \textbf{0.476} & 0.836 \\
  Fine-tuned & 0.239          & \textbf{0.190} & \textbf{1.000} & \textbf{0.959} & 0.459          & \textbf{0.999} \\
  \midrule
  Piezo-only & 0.482 & 0.117 & 0.998 & 0.692 & 0.630 & 0.736 \\
  \bottomrule
 \end{tabular}
 We show accuracies for each class, split over the competition days and model variants.
 Accuracies reflect the ratio of correctly identified chunks.
 \end{threeparttable}
\end{table} 

We compare two model variants that differ in the data used during training.
First, we propose a general variant trained using the entire dataset and the threshold for distinguishing contact set to $\SI{-26}{\dBFS}$, slightly exceeding the noise floor of the piezo microphone.
\cref{tab:confusion} shows the confusion matrices of the general model variant for both the test set and the competition runs.
Please note that we explicitly tuned the model to produce low false-positive rates.
Due to the vibration motor inside the actuator being slow in its response relative to the prediction rate, and the vibration intensity of rough classification results set relatively high, even misclassifications of single chunks can give the operator the false impression of sensing a rough surface.
On the other hand, the correct classification of only several chunks suffices to convey the desired impression when sliding the finger over a rough texture.

Second, we evaluate a fine-tuned model variant optimized for participation in the competition, using a reduced set of training objects including samples of the stones encountered during the competition runs (\cref{fig:textures}).
Here, we set the threshold for distinguishing contact dynamically to $\SI{50}{\percent}$ of the RMS loudness of all files with the respective label.
This increases the amount of \textit{non-valid} classification results when applying light pressure onto an object at the benefit of further decreasing the false-positive rate. 
\cref{tab:variants} compares the classification accuracy of both model variants during the competition runs and for the test set.
While the accuracy of classifying smooth objects is very high for both model variants, the fine-tuned variant substantially outperforms the general one here, but falls slightly short w.r.t. rough objects.

Furthermore, to justify the usage of the additional MEMS microphone, we show the accuracy of the general model variant using only the piezo microphone during training and inference.
The low accuracy for classifying smooth textures and the associated false-positive rates of $\SI{3.3}{\percent}$ for the competition runs and $\SI{15.5}{\percent}$ for the test set are insufficient for our application.

\subsection{ANA Avatar XPRIZE Competition}

\begin{figure}
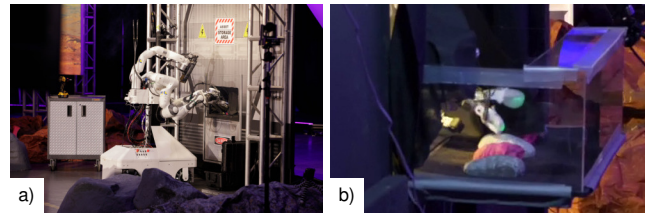

     \centering
     \newlength{\taskheight}\setlength{\taskheight}{.32\linewidth}
     \begin{tikzpicture}[a/.style={inner sep=0pt}, l/.style={anchor=south west, fill=white, font=\sffamily\scriptsize}]
        \node[a] (img1) {\includegraphics[height=\taskheight]{images/task/task_1.jpg}};
        \node[a, right=.1cm of img1] (img2) {\includegraphics[height=\taskheight,clip,trim=470px 40px 0px 250px]{images/task/task_2.png}};
        \node[l] at ($(img1.south west)-(0.01,0.01)$) {a)};
        \node[l] at ($(img2.south west)-(0.01,0.01)$) {b)};
     \end{tikzpicture}
     \caption{Our avatar robot during the roughness sensing and stone retrieval task. a) Approaching and reaching through the box opening covered by a curtain. b) Sensing one of the stones inside the box with the instrumented finger.}
     \label{fig:task}
\end{figure}

\begin{figure}
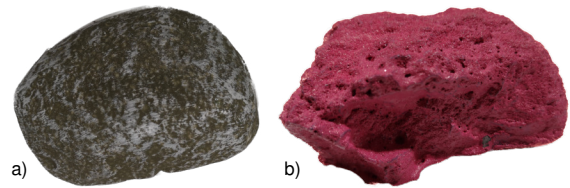

     \centering
     \newlength{\stonesheight}\setlength{\stonesheight}{.28\linewidth}
     \begin{tikzpicture}[a/.style={inner sep=0pt}, l/.style={anchor=south west, font=\sffamily\scriptsize}]
        \node[a] (img1) {\includegraphics[height=\stonesheight]{images/textures/smooth_texture.png}};
        \node[a, right=.2cm of img1] (img2) {\includegraphics[height=\stonesheight]{images/textures/rough_texture.png}};
        \node[l] at ($(img1.south west)-(0.1,0.01)$) {a)};
        \node[l] at ($(img2.south west)-(0.01,0.01)$) {b)};
     \end{tikzpicture}
     \caption{Samples of the (a) smooth and (b) rough stone textures encountered during the roughness sensing and stone retrieval task in the competition.}
     \label{fig:textures}
\end{figure}

\cref{fig:task} shows our avatar robot during the last task in the finals testing event of the ANA Avatar XPRIZE competition.
During this task, the operator was required to find and retrieve one of the rough stones, purely based on their haptic perception.
In particular, there were five stones lined up on an anti-slip mat in a small box, with an opening that blocked the operator's vision through a curtain.
Three of the stones had a smooth texture, while two had a rough texture and were highlighted in pink color, which was only relevant for the audience to distinguish the stones.
\cref{fig:textures} shows a close-up of sample textures encountered in the competition.

\begin{figure}
     \includegraphics[width=1.0\columnwidth]{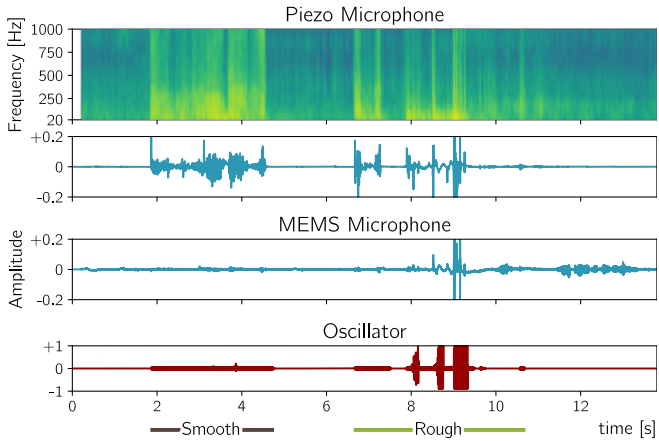}
     \caption{Microphone and oscillator signals during the roughness sensing and stone retrieval task on Day~1.
     Ground truth times of contact and rock type are shown at the bottom.}
     \label{fig:audio_day1}
\end{figure}

\begin{figure}
     \includegraphics[width=1.0\columnwidth]{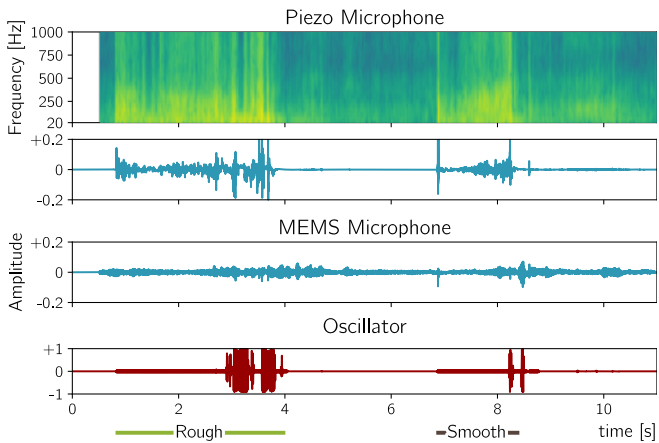}
     \caption{Microphone and oscillator signals during the roughness sensing and stone retrieval task on Day~2.
     Ground truth times of contact and rock type are shown at the bottom.
     Note the false positive classification at the end of the smooth stone, where the finger
     slipped off and hit the surface underneath. The operator correctly interpreted this as the edge of the stone.}
     \label{fig:audio_day2}
\end{figure}

In total, the task was encountered up to three times per team during the event.
Each time, a different operator judge was controlling our avatar robot.
The operators were members of the XPRIZE jury and impartial in their judgment of task completion.
They were trained for $\SI{45}{\min}$, directly before the run, to familiarize themselves with the system.
However, only a fraction of this time was allocated to training for this specific task, as nine previous tasks needed to be completed to advance to the final task.
All three task attempts were successful.
\cref{fig:audio_day1} and \cref{fig:audio_day2} show the measured audio and generated feedback signals of Days 1 \& 2, respectively.

\pagebreak

\begin{table}[t]
     \centering
	\begin{threeparttable}
	\caption{Task completion times.}
	\label{tab:times}
	\begin{tabular}{ ccccc }
		\toprule
          & \textbf{NimbRo} & Pollen Robotics & Northeastern & Avatrina \\
		\midrule
          Day~1 &
          \textbf{$\mbox{1:06}$}&
          {$\mbox{2:24}$}&
          N/A&
          {$\mbox{4:48}$}\\
          Day~2 &
          \textbf{$\mbox{1:02}$}&
          {$\mbox{1:59}$}&
          {$\mbox{9:27}$}&
          N/A\\
		\bottomrule
	\end{tabular}
	Time is given in min:sec and includes roughness sensing and stone retrieval.
     N/A: not attempted.
	\end{threeparttable}
\end{table}

While the first run on Qualification Day was not public and results are not available, the runs on Testing Days 1 \& 2 were broadcasted by the organizers\footnote{\url{https://www.youtube.com/watch?v=EmESa2Olq4c}}, allowing for a comparison with all other teams that completed the task.
\cref{tab:times} shows that we completed the task considerably faster than any other team.
However, it is worth mentioning that other factors besides the haptic feedback may have contributed to the reported times. %

\section{Conclusion}

We presented an audio-based haptic teleperception system that is based on low-cost, compact components.
The system was proven to be very effective at the ANA Avatar XPRIZE competition finals, winning the first prize.
Even though the system was mostly trained and tested on stone surfaces, the method can be adapted easily to other surface kinds by collecting the appropriate training data.

\printbibliography

\end{document}